\documentclass[11pt,a4paper]{article}
\usepackage{authblk}
\usepackage[hyperref]{acl2021}
\usepackage{times}
\usepackage{latexsym}

\usepackage{enumitem}
\usepackage{soul}
\usepackage{url}
\usepackage{graphicx}
\usepackage{amsmath}
\usepackage{multirow}
\usepackage{amsthm}
\usepackage{booktabs}
\usepackage[noend]{algpseudocode}
\usepackage{algorithmicx}
\usepackage{algorithm}
\usepackage{IEEEtrantools}
\usepackage{amssymb, amscd, amsfonts}
\usepackage{arydshln}
\usepackage{makecell}
\usepackage{xcolor}
\usepackage{caption}
\usepackage{subcaption}
\usepackage[export]{adjustbox}

\newcommand{\tabincell}[2]{\begin{tabular}{@{}#1@{}}#2\end{tabular}}

\usepackage{microtype}

\usepackage{xspace}

\aclfinalcopy 
\def\aclpaperid{2534} 

\newcommand{\mymodel}{\textsc{PRobr}\xspace}
\newcommand{\prover}{\textsc{PRover}\xspace}
\title{Probabilistic Graph Reasoning for Natural Proof Generation}

\author[$\dagger$*]{\bf Changzhi Sun}
\author[$\dagger$*]{\bf Xinbo Zhang}
\author[$\dagger \mathsection$]{\bf Jiangjie Chen}
\author[$\dagger \mathparagraph$]{\bf Chun Gan}
\author[$\ddagger$]{\\ \bf Yuanbin Wu}
\author[$\dagger$]{\bf Jiaze Chen}
\author[$\dagger$]{\bf Hao Zhou}
\author[$\dagger$]{\bf Lei Li}
\affil[$\dagger$]{ByteDance AI Lab}
\affil[$\mathsection$]{School of Computer Science, Fudan University}
\affil[$\mathparagraph$]{Math Department, University of Wisconsin--Madison}
\affil[$\ddagger$]{School of Computer Science and Technology, East China Normal University}
\affil[  ]{\tt \{sunchangzhi,zhangxinbo.freya\}@bytedance.com}
\affil[  ]{\tt \{chenjiaze,zhouhao.nlp,lileilab\}@bytedance.com}
\affil[  ]{\tt jjchen19@fudan.edu.cn, cgan5@wisc.edu, ybwu@cs.ecnu.edu.cn}

\begin{document}

\maketitle

\renewcommand{\thefootnote}{\fnsymbol{footnote}}
\footnotetext[1]{Equal contribution.}
\renewcommand{\thefootnote}{\arabic{footnote}}

\begin{abstract}
In this paper, we investigate the problem of reasoning over natural language statements.
Prior neural based approaches do not explicitly consider the inter-dependency among answers and their proofs.
In this paper, we propose \mymodel,
a novel approach for joint answer prediction and proof generation.
\mymodel defines a joint probabilistic distribution over all possible proof graphs and answers via an induced graphical model. 
We then optimize the model using variational approximation on top of neural textual representation. 
Experiments on multiple datasets under diverse settings (fully supervised, few-shot and zero-shot evaluation) verify the effectiveness of \mymodel, 
e.g., achieving 10\%-30\% improvement on QA accuracy in  few/zero-shot evaluation.
Our codes and models can be found at \url{https://github.com/changzhisun/PRobr/}.
\end{abstract}

\section{Introduction}
\label{sec:intro}

Automatic reasoning over explicitly provided knowledge has been a persistent goal of AI ~\cite{newell1956logic,mccarthy1960programs}. 
Early approaches focus on reasoning over formal (logical or probabilistic) representations.
However, automatically constructing and reasoning over formal representations remain challenging.
To bypass these challenges, in this work, we investigate reasoning over natural language statements instead of formal representations.

\begin{figure}[t]
    \centering
        \includegraphics[width=3.0in]{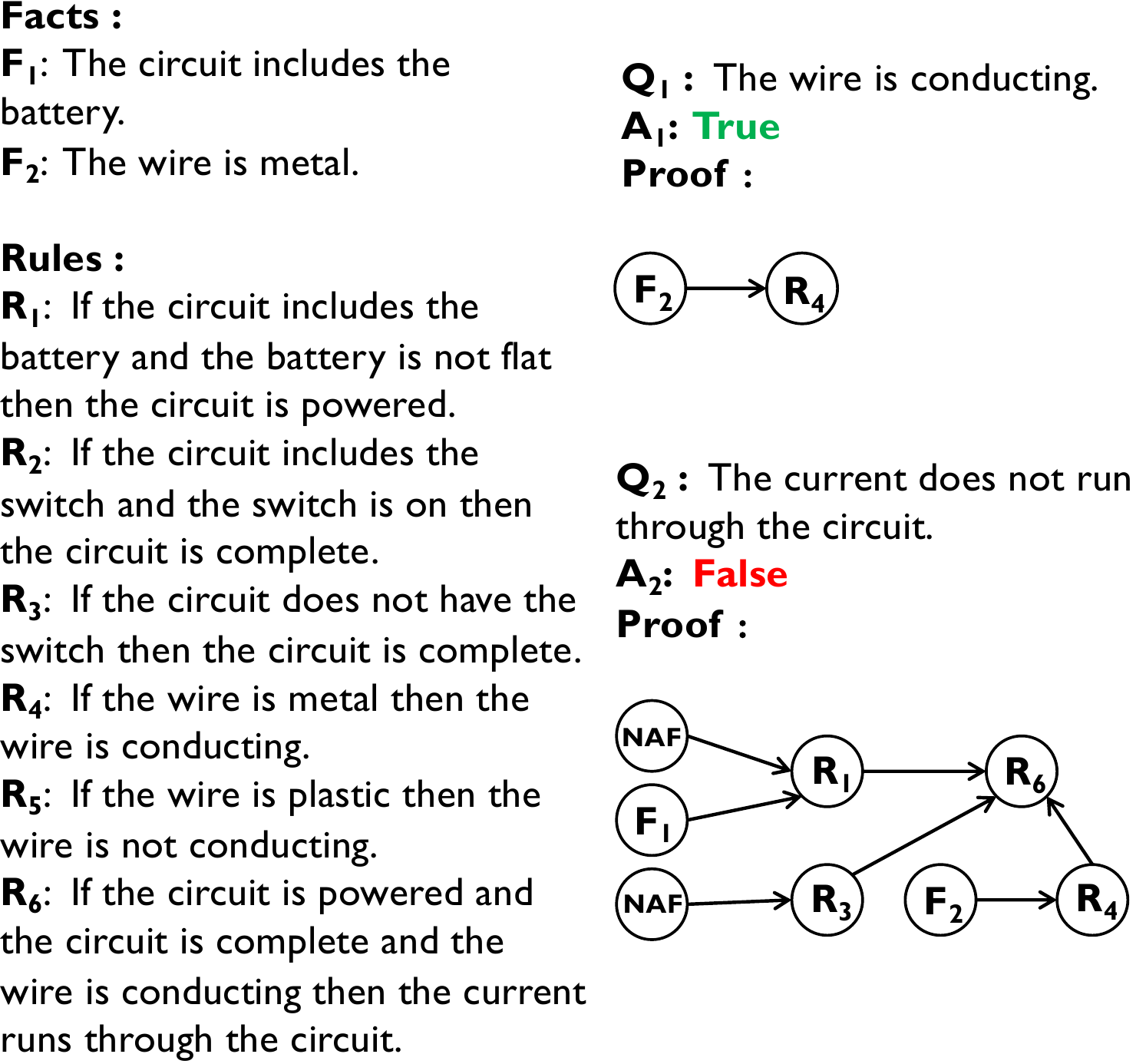}
    \caption{An example of reasoning over natural language statements. The goal is to predict the answer (true/false) and generate the proof graph.}
    \label{fig-example}
\end{figure}

Given a set of facts and rules and a query (expressed in natural language),
we aim to predict the answer and provide proof to prove or disprove the query.
For example, in Figure~\ref{fig-example},
there are two facts, six rules and two queries, each of which is expressed by natural language.
To predict the true/false of each query, starting from the facts, we need to reason deductively by applying given rules until we can derive the truth value of the query.
The process of deduction can be represented as a graph,
whose node is either a fact, rule or special NAF node (explained in the Section \ref{sec:semantics}).
Generating answer and proof together makes a system easier to interpret and diagnose. 


Recent work by \prover ~\cite{saha2020prover} first explored this problem through two modules: question answering and proof generation.
It trains these two modules through implicit parameter sharing,
and then uses integer linear programming (ILP) to enforce consistency constraints (only test time).
It is difficult to ensure that the proof generation module contributes to the question answering module,
because the proof is not explicitly involved in the answer prediction.
Parameter sharing becomes more limited under few/zero-shot settings, as demonstrated in our experiments.
We expect the proof to enhance the capability of question answering, especially under few/zero-shot settings. 
One promising solution is to explicitly exploit more interaction between question answering and proof generation.


In this paper, we propose \mymodel, a novel \textbf{prob}abilistic graph \textbf{r}easoning framework for joint question answering and proof generation.
\mymodel defines a joint distribution over all possible proof graphs and answers with an undirected \emph{probabilistic graphical model} (PGM).
It directly characterizes the interaction between proofs and answers.
PGMs generally incur intractable learning and inference for the complex graph
 ~\cite{koller2009probabilistic}.
For example, 
computing normalization constant in PGMs using traditional probabilistic propagation algorithm (e.g.sum-product algorithm \cite{910572}) requires large time complexity.
Therefore, we propose a variational approach to maximize the pseudolikelihood of joint distribution to optimize the model more efficiently.
First,
a variational distribution was introduced based on mean-field assumption.
Then we maximize the pseudolikelihood of joint distribution given the output of variational distribution.
At the same time, we align these two distributions using the training data.
\mymodel can be efficiently trained by stochastic gradient descent. 
Our contributions are summarized as follows\footnote{Our codes and models can be found at \url{https://github.com/changzhisun/PRobr/}.}:
\begin{itemize}[leftmargin=*, itemindent=1pc]
    \item We propose \mymodel for joint question answering and proof generation,
    which defines a joint distribution over all possible proofs and answers with an undirected PGM to capture more dependencies.
    \item We present an efficient variational approximation method to learn \mymodel.
    \item  Experiments on several datasets verify the effectiveness of \mymodel under  multiple settings (supervised, few-shot, and zero-shot evaluation).
\end{itemize}

\section{Task Definition}
\label{sec:prelim}

To reason over natural language statements, 
we design to answer the query and generate corresponding proof generation jointly.
Figure \ref{fig-example} shows an example.
Given a declarative query $Q$, and given relevant facts and rules (expressed in natural language), the task aims to predict the answer $A$ (true/false) to the query $Q$ based on the closed-world assumption (described in \ref{sec:semantics}). Meanwhile, it generates a proof $P$ (described in \ref{sec:proof}) to prove or disprove $Q$. 


\begin{figure*}[htb]
\centering
    \begin{subfigure}[b]{0.47\linewidth}
        \includegraphics[width=\linewidth]{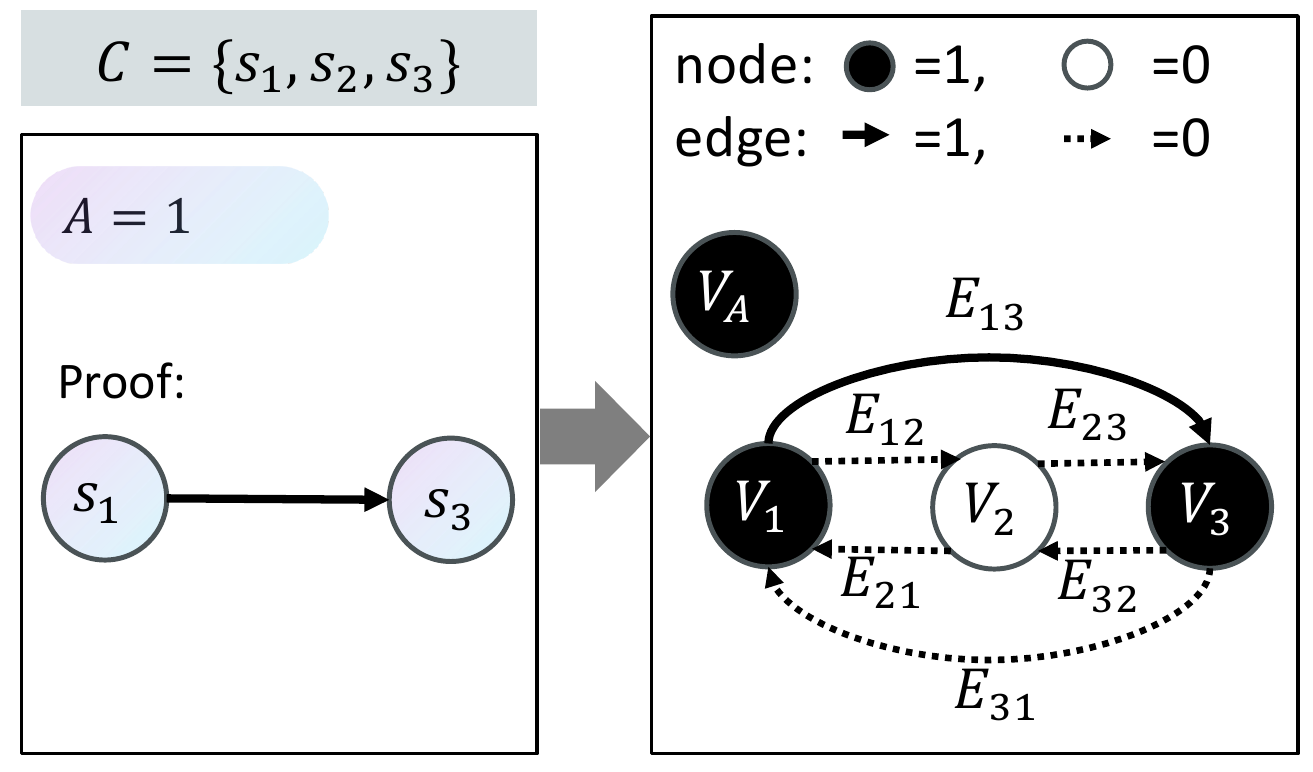}
        \caption{Proof graph and its induced random variables.}
        \label{fig-overiew-new:a}
    \end{subfigure}%
    \begin{subfigure}[b]{0.52\linewidth}
        \includegraphics[width=\linewidth]{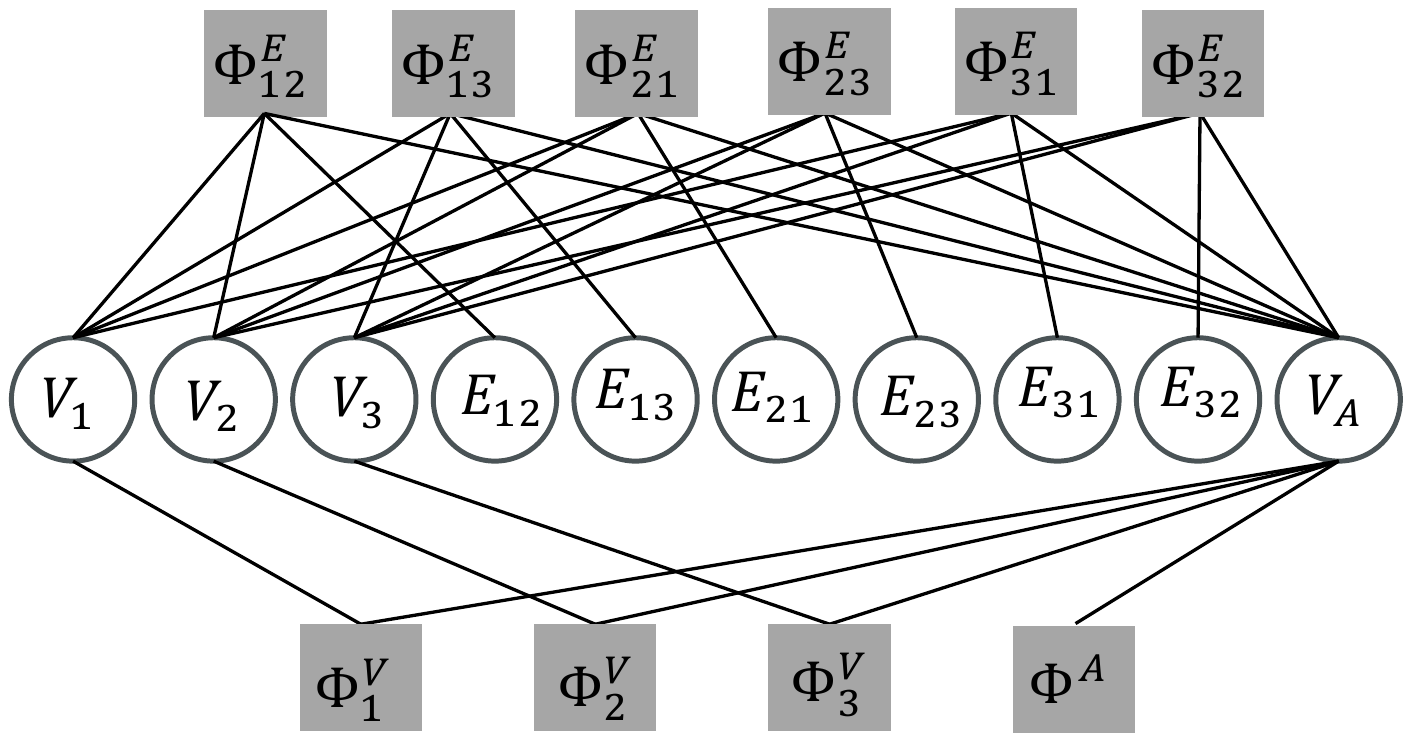}
        \caption{Factor graph induced by the proof graph.}
        \label{fig-overiew-new:b}
    \end{subfigure}
    \caption{Joint probabilistic distribution by assigning indicator variables for the answer and proof. The solid circles and lines in \ref{fig-overiew-new:a} indicate that these corresponding statements and edges are in the final proof graph. }
    \label{fig:overview}
\end{figure*}

    
\subsection{Semantics}
\label{sec:semantics}
We adopt the semantics of Datalog \cite{ceri1989you} in this work. Following prior work \cite{clark2020transformers},
we make a closed-world assumption (CWA), which means a fact is true if it can be deduced based on a given context, and any fact not provable is assumed false.
And we use negation as failure (NAF) \cite{Clark1978}, a rule of inference which allows one to deduce that $\mathtt{NOT}$ $\mathtt{S}$ is true if all possible proofs of a statement $\mathtt{S}$ fail. 
For example, in Figure \ref{fig-example}, the \texttt{NAF} node before $\mathtt{R_3}$ represents ``the circuit does not have the switch''.
Note that under this semantics,
negative facts and negative rules are not allowed because of redundancy under the CWA assumption.

\subsection{Formulations} 
\label{sec:proof}
A proof is a directed acyclic graph (Figure \ref{fig-example}).
Each node is either a fact, rule
or special \texttt{NAF} node.
Each edge directs from either a fact (or \texttt{NAF}) to a rule or a rule to another rule, which indicates that a fact is consumed by a rule, or another rule consumes a rule, respectively.
For simplicity, let context $C = \{s_1, \dots, s_n\}$ denote the collection of sentence, each of which is a fact or rule.

\paragraph{Proof Formulation}
We assign an indicator variable (0/1) for each possible node and edge, to vectorize the structure of a given proof $P$. Specifically, we introduce the indicator variables $V = \{V_i\}_{i=1}^n$ for each element $s_i$ in the context $C$, and an indicator variable $ E = \{E_{ij} \}_{i, j = 1}^n$ ($i \neq j$) for a possible edge connecting from node $s_i$ to node $s_j$, where: 
\begin{itemize}[leftmargin=*, itemindent=1pc]
    \item $V_i = 1$ indicates $s_i$ is in the proof $P$, while $V_i = 0$ means $s_i$ is absent. 
    \item $E_{ij} = 1 $ indicates there is an edge directing from $s_i$ to $s_j$, while $E_{ij} = 0 $ means $s_i$ cannot direct to $s_j$ in the proof by an edge.
\end{itemize}

In addition, we assign a binary answer variable $A$ to indicate the true value of the query.
Figure \ref{fig-overiew-new:a} shows a simplified example, where context $C = \{s_1, s_2, s_3\}$, and the query can be decided as true by a very simple proof, consisting of only two nodes ($s_1$ and $s_3$) and a single edge (from $s_1$ to $s_3$). The proof can be represented by the following variables: $A = V_1 = V_3 = E_{13} = 1, V_2 = E_{12} = E_{21} = E_{23}= E_{31} = E_{32}=0$.

\section{Approach}
\label{sec:approach}

We introduce the proposed framework \mymodel, which jointly provides the answer to the given query over natural contexts and generates corresponding proof.
Different from \prover that makes independence assumption,
\mymodel can capture more dependencies between the proof and answer.
\mymodel defines a joint distribution over all possible proofs and answers with an undirected graphical model (Section \ref{sec-prob-formulation}),
and we use neural networks to parameterize each component (Section \ref{sec-param}).
To optimize \mymodel efficiently, 
we adopt a variational approach to maximize the pseudolikelihood of joint distribution (Section \ref{sec-optim}).
Finally, we introduce the strategy during inference (Section \ref{sec-prediction}).

\subsection{Overview}
\label{sec-prob-formulation}
We start by formalizing joint question-answering module and proof-generation module in a probabilistic way.
We clarify some notations as follows:
\begin{itemize}[leftmargin=*, itemindent=1pc]
    \item A context $C = \{s_1, \dots, s_n\}$, $s_i$ is a sentence.
    \item A query $Q$.
    \item Answer variable $A$, it can take any value $a$ in $\{0, 1\}$.
    \item Node variables $\mathcal{V} = \{V_i\}_{i=1}^n$, each $V_i$ can take any value $v_i$ in $\{0, 1\}$. 
    \item Edge variables $ \mathcal{E} = \{E_{ij} \}_{i, j = 1}^n$ ($i \neq j$), each $E_{ij}$ can take any value $e_{ij}$ in $\{0, 1\}$.
    \item Let $Y \triangleq (A, \mathcal{E}, \mathcal{V})$ denote all output variables.
\end{itemize}
In our notation, we use uppercase letters for variables (e.g., $Y, A, V_i, E_{ij}$) and lowercase letters for variables that take values (e.g., $y, a, v_i, e_{ij}$).

Given a context $C$ and a query $Q$,
\mymodel tries to assign true/false values for all variables,
including answer variable $A$,  node variables $V$ and
edge variables $E$.
We define a joint distribution over all possible $Y$, officially denoted as $p(Y)$:~\footnote{We drop the input variables for clarity.} 
\begin{IEEEeqnarray}{rl}
\small
    p(Y &= y) \propto  \nonumber\\
    &\Phi^A(a) 
     \prod_{i} \Phi_i^V ( v_i, a ) 
    \prod_{i, j} \Phi_{ij}^E(v_i, v_j, e_{ij}, a) ~ ~ ~ 
    \label{eq-joint}
\end{IEEEeqnarray}
Different from \prover that makes independent assumption,
such a factorization of Equation \ref{eq-joint} can characterize the interaction between the variables  $V_i, V_j, E_{ij}$ and $A$.
Figure \ref{fig-overiew-new:b} shows the factor graph of joint distribution $p(Y)$ for the example in Figure \ref{fig-overiew-new:a}.
Theoretically, when we have the ground truth $y^*$, \footnote{We use $*$ to indicate the ground truth in the text.} we can minimize the following objective:
\begin{IEEEeqnarray}{rl}
    \mathcal{L}_\mathrm{joint} = - \log p(Y = y^*)
    \label{eq-joint-loss}
\end{IEEEeqnarray}
However, the normalization constant of $p(Y)$ is hard to calculate due to high-order factors of large size (RHS of Equation \ref{eq-joint}).
In this paper, we provide a variational-based solution to optimize objective $\mathcal{L}_\mathrm{joint}$ (Section \ref{sec-optim}).

\subsection{Parameterization}
\label{sec-param}
We use neural networks to parameterize each potential function of Equation \ref{eq-joint}: $\Phi^A, \Phi_i^V$ and $\Phi_{ij}^E$.
\paragraph{Text Representation Network} Given a context $C$ and a query $Q$, to obtain a contextual representations,
we use RoBERTa \cite{liu2019roberta} as our backbone network following \cite{clark2020transformers,saha2020prover}.
The input to RoBERTa is the concatenation of $C$ and $Q$, separated by $[\texttt{SEP}]$ tokens, denoted as:
$[\texttt{CLS}], C, [\texttt{SEP}], [\texttt{SEP}], Q, [\texttt{SEP}]$.

\paragraph{Potential Function for the Answer ($\Phi^A$)}
After the RoBERTa encoding, we can get the global representation of the entire input through the first token $[\texttt{CLS}]$, denoted as $h_{[\texttt{CLS}]}$.
To score the possible values of variable $A$, i.e. 0 or 1, 
we use a multi-layer perceptron (MLP) as a nonlinear transformation:
\begin{IEEEeqnarray*}{c}
    \left [ \begin{array}{c}
        \Phi^A(A = 0)  \\
        \Phi^A(A = 1)
    \end{array}\right ] = 
    \mathrm{MLP}_1(h_{[\texttt{CLS}]}) \in \mathbb{R}^2 
\end{IEEEeqnarray*}

\paragraph{Potential Function for Statements ($\Phi_i^V$)}
For each sentence $s_i$ (a fact or a rule),
we compute the sentence representation $h_{s_i}$ by performing a mean pool of the all token representation based on the output of RoBERTa.
It is worth noting that \texttt{NAF} is a special fact, 
we calculate $h_{\texttt{NAF}}$ through linear transformation on $h_{\texttt{CLS}}$.
To score the possible values of variables $(V_i, A)$,
we also use another MLP as a score function:
\begin{IEEEeqnarray*}{c}
    \left [ \begin{array}{c}
        \Phi^V_i(V_i = 0, A = 0)  \\
        \Phi^V_i(V_i = 0, A = 0)\\
        \Phi^V_i(V_i = 0, A = 0) \\
        \Phi^V_i(V_i = 1, A = 1) \\
    \end{array}\right ] = 
    \mathrm{MLP}_2(h_{s_i}) \in \mathbb{R}^{4}, 
\end{IEEEeqnarray*}
where the dimension $4$ indicates the number of possible values for the combination of variables $V_i$ and $A$.
We share the parameters of $\mathrm{MLP}_2$ across all sentences.

\paragraph{Potential Function for Statement Relations ($\Phi_{ij}^E$)}
For each sentence pair $(s_i, s_j)$,
we obtain the sentence pair representation $h_{s_i, s_j}$,
by concatenating $h_{s_i}$ and $h_{s_j}$ with their element-wise difference (directionality).
To score four variables $(V_i, V_j, E_{ij}, A)$ simultaneously,
similarly, we use a new MLP as score function:
\begin{IEEEeqnarray*}{c}
     \left [ \begin{array}{c}
        \Phi^E_{ij}
        \left ( \begin{array}{c}
            V_i = 0,  V_j  = 0,\\
            E_{ij} = 0,  A = 0 
        \end{array}\right)  \\
        \vdots \\
        \Phi^E_{ij}\left ( \begin{array}{c}
            V_i = 1,
            V_j  = 1,\\
            E_{ij} = 1,
            A = 1 
        \end{array}\right) \\
    \end{array}\right ] \begin{array}{l}
        =  \mathrm{MLP}_3(h_{s_i, s_j})   \\
        ~ ~ ~ \in \mathbb{R}^{16},
    \end{array}
      \\
      h_{s_i, s_j} = h_{s_i} \oplus h_{s_j} \oplus (h_{s_i} - h_{s_j}),
 \end{IEEEeqnarray*}
where $\oplus$ is the vector concatenation, and 
the dimension $16$ indicates the number of possible values for the combination of four variables $(V_i, V_j, E_{ij}, A)$.
We also share the parameters of $\mathrm{MLP}_3$ across all sentence pairs.

\subsection{Learning the Model}
\label{sec-optim}
To tackle the challenge of optimizing $\mathcal{L}_\mathrm{joint}$ (Equation \ref{eq-joint-loss}),
we adopt the widely used pseudolikelihood as an alternative objective for optimization \cite{richardson2006markov},
bypassing the calculation of the normalization constant.
\paragraph{Pseudolikelihood}
Given a set of variable $Y$, the pseudolikelihood of $Y$ is defined as:
\begin{IEEEeqnarray*}{c}
    p_\mathrm{pseduo}(Y) = 
    \prod_{y \in Y} p(y | Y_{-y}) = \\
    p(A|\mathcal{E}, \mathcal{V}) \prod_{i} p(V_i | Y_{-V_i}) \prod_{i,j}p(E_{ij} | Y_{-E_{ij}})
    \label{eq-pseudo}
\end{IEEEeqnarray*}
When we have the ground truth $y^*$,
we can minimize the following objective:
\begin{IEEEeqnarray*}{rl}
    \mathcal{L}_\mathrm{pseudo} = -\log p_\mathrm{pseudo}(Y = y^*)
\end{IEEEeqnarray*}
However,
it is difficult to decode the optimal assignments based on the  pseudolikelihood (Equation \ref{eq-pseudo}).
There is a rich body of literature on how to decoding in a sampling way (Chapter 12 \cite{goodfellow2016deep}).
In this paper, however, we choose a modern approach using variational approximation.

\paragraph{Variational Approximation}
We approximate pseudolikelihood of $Y$ with a mean-field \cite{opper2001advanced} variational distribution $q(Y)$,
in which $y \in Y$ is independent of each other.
Similarly, we parameterize each independent distribution with a neural network.
Formally, $q(Y)$ is formulated as below:
\begin{IEEEeqnarray*}{cl}
    q(Y) &= q(A) \prod_{i} q(V_i) \prod_{i, j} q(E_{ij}) \\
    q(A) &= \mathrm{Softmax}(\mathrm{MLP}_4(h_\texttt{CLS}))  \in \mathbb{R}^2\\
    q(V_i) &= \mathrm{Softmax}(\mathrm{MLP}_5(h_{s_i})) \in \mathbb{R}^2 ,\\
    q(E_{ij}) &= \mathrm{Softmax}(\mathrm{MLP}_6(h_{s_i, s_j})) \in \mathbb{R}^2 .
    \label{eq-variation}
\end{IEEEeqnarray*}
Once the variational distribution $q(Y)$ is obtained, it can provide conditions for pseudolikelihood $p(y|Y_{-y})$, thus avoiding the sampling process to obtain the optimal assignments.
In the optimization process,
we adopt the simple strategy to update the parameters of $p$ and $q$.
\begin{itemize}[leftmargin=*, itemindent=1pc]
    \item For node and edge variables, we optimize
    \begin{IEEEeqnarray*}{rl}
        \mathcal{L}_\mathrm{node} &= -\sum_i \log q(V_i = v_i^*), \\
        \mathcal{L}_\mathrm{edge} &= -\sum_{i, j} \log q(E_{ij} = e_{ij}^*).
    \end{IEEEeqnarray*}
    \item For answer variable, we optimize
    \begin{IEEEeqnarray*}{rl}
    \mathcal{L}_\mathrm{qa} &= -\log p(A = a^* | \hat{\mathcal{E}}, \hat{\mathcal{V}}),
    \end{IEEEeqnarray*}
    where $\hat{\mathcal{E}} = \{\hat{e}_{ij}\}, \hat{\mathcal{V}} = \{ \hat{v}_i\}$ is the predictions of variational model \footnote{We also experiment with the gold proof to optimize $\mathcal{L}_\mathrm{qa}$ (Section \ref{sec:abla}).}. 
\end{itemize}
The final objective is to minimize:
\begin{IEEEeqnarray*}{rl}
\mathcal{L}_\mathrm{final} &= \mathcal{L}_\mathrm{qa} + \mathcal{L}_\mathrm{node} + \mathcal{L}_\mathrm{edge}
\end{IEEEeqnarray*}
Overall, \mymodel is a mixture of independent (variational) model and undirected graphical model through some reasonable approximations. 
Our final optimized distribution can be decomposed as directed graphical model $q(\mathcal{V}) q(\mathcal{E}) p(A| \mathcal{E}, \mathcal{V})$, where $q(V), q(E)$ adopts the independent factorized probability, and $p(A| \mathcal{E}, \mathcal{V})$ is implied by the undirected graphical model (Equation \ref{eq-joint}).
In this way, \mymodel enjoys  the advantage of global normalization (undirected graphical model) and is easier to optimize (directed graphical model).

\paragraph{Discussion}
Another way to achieve consensus between $q(Y)$ and $p_\mathrm{pseudo}(Y)$ is to directly optimize the KL divergence:
\begin{IEEEeqnarray*}{rl}
    \mathcal{L}_\mathrm{kl} = \sum_{y \in Y} \mathrm{KL}\left(q(y) || p(y| Y_{-y})\right)
\end{IEEEeqnarray*}
However, $\mathcal{L}_\mathrm{kl}$ does not bring any improvement for supervised learning (Section \ref{sec:abla}), hence we exclude it during training. \mymodel can be easily extended to semi-supervised learning scenario by using this $\mathcal{L}_\mathrm{kl}$ term. Specifically, minimize the $\mathcal{L}_\mathrm{final}$ for the labeled data; and minimize the $\mathcal{L}_\mathrm{kl}$ for the unlabeled data. We save this for future work.

\subsection{Inference}
\label{sec-prediction}
After training,
for nodes and edges, we choose the predictions of the variational model, 
and for answers, we choose the prediction of the joint model based on the output of variational model.
In addition, we also employ the Integer Linear Programming (ILP) to enforce consistency constraints following \cite{saha2020prover}.

\section{Experiments}
\label{sec:exp}

To evaluate the effectiveness and generality of our \mymodel model, we conduct both fully supervised learning, few-shot learning, and zero-shot learning over several datasets\footnote{\url{https://allenai.org/data/ruletaker}}  against two baselines: RuleTakers and \prover\footnote{Please refer to supplementary materials for our hyperparameter and computing infrastructure. For RuleTakers and \prover, we directly adopt results reported in their papers if exist, and for extra setting beyond papers, we reproduce the baselines using provided codes and parameters:\url{https://github.com/swarnaHub/PRover}.}. 
\subsection{Datasets and Metrics}
We use three datasets (DU0-DU5, Birds-Electricity, ParaRules) introduced by \cite{clark2020transformers}.

\paragraph{DU0-DU5} 
DU$d$ ($d$=0,1,2,3,5) are five synthetic datasets, each containing 100k queries with theories expressed in templated English, proof graphs expressed in natural language, and answers described as \emph{True/False}. Answers require reasoning up to depth $d$ for queries in DU$d$.

\paragraph{Birds-Electricity}
This dataset is a test-only dataset of 5k samples in total. It describes birds and electric circuit, which was used to evaluate the out-of-distribution performance of the models.

\paragraph{ParaRules}
ParaRules is a dataset generated and paraphrased from sampled theories (facts + rules). It contains 40k queries against $\approx$2k theories, where the original templated English facts and rules are creatively paraphrased into more diverse natural language by crowdsourcing. For example, the fact ``Dave is cold'' can be rephrased as``After Dave got wet in the rain, he feels cold''; the rule ``If someone is nice then they are young'' can be rephrased into ``A person described as being nice will certainly be young''. Different from 
DU$d$ and Birds-Electricity dataset composed of synthetic language, ParaRules can better test models' reasoning ability over human-like language.

\paragraph{Metrics}
We evaluate the performance considering both answers and proofs. For answers, we evaluate the \textbf{QA Accuracy} (\textbf{QA}). For proofs, we evaluate the \textbf{Proof Accuracy} (\textbf{PA}), and PA refers to the fraction of examples where generated proof matches exactly with the gold proof. We also report \textbf{Full Accuracy} (\textbf{FA}) to denote the faction of examples where both the answer and the proof are exactly correct.

\subsection{Fully Supervised Learning}
\label{sec:fully-sup}
For the supervised setting, we train \mymodel on the training split of the DU5 dataset with gold answer and gold proof and evaluate on the test split of DU5. We evaluate above metrics of varying depths $d$ against two state-of-the-art baselines: RuleTakers \cite{clark2020transformers} and \prover \cite{saha2020prover}, showed in Table \ref{tab:full-supervised}. For RuleTakers and \prover, we directly adopt the results reported in their paper. Note that RuleTakers can not generate a proof, so we only report the PA and FA on \prover and \mymodel. The corresponding validation set results can be found in the supplementary materials . 

Overall, at each depth, \mymodel generates comparable or superior QA accuracy to baselines. And for 88.8\% of test examples, \mymodel can generate exact proofs and answers. 
Similar to \prover, the full accuracy matches the proof accuracy for \mymodel, showing that in this fully supervised setting, full accuracy depends on proof accuracy at each depth. The predicted answer is always correct when the corresponding proof is correct. Actually, answering predicting is much easier than a proof generation. 

When increasing depth, \mymodel provides accurate answers without any loss in QA performance. It becomes harder to generate correct proofs for both \prover and \mymodel, while \mymodel outperforms \prover by 7 points of proof accuracy ($65.1\% \rightarrow 72.2\%$) at depth 5.


\begin{table}[t]
    \centering

    \small
    \scalebox{0.9}{
        \begin{tabular}{p{0.2cm}p{0.6cm}p{0.4cm}cccccc}
            \toprule
            \multirow{2}{*}{\textbf{D}} & \multirow{2}{*}{\textbf{Cnt}}&\multicolumn{3}{c}{\textbf{QA}}& \multicolumn{2}{c}{\textbf{PA}}&\multicolumn{2}{c}{\textbf{FA}}\\
		\cmidrule{3-5}\cmidrule{6-7}\cmidrule{8-9}
            &&RT&PV&PB& PV&PB&PV&PB  \\
                
            \midrule
            \textbf{0}      &6299      &100    &100 &\textbf{100}&98.4&\textbf{98.4}&98.4&\textbf{98.4} \\
            \textbf{1}    &4434     &  98.4    & 99.0 &\textbf{99.9}&93.2&\textbf{94.3}&93.1&\textbf{94.3}   \\
            \textbf{2}   &2915       &  98.4     &98.8 &\textbf{99.9}&84.8&\textbf{86.1}&84.8&\textbf{86.1}  \\
            \textbf{3}   &2396       & 98.8      & 99.1 &\textbf{100}&80.5&\textbf{82}&80.5&\textbf{82} \\
            \textbf{4}   &2134       &  99.2     & 98.8 &\textbf{100}&72.5&\textbf{76.1}&72.4&\textbf{76.1} \\
            \textbf{5}   &2003       & 99.8      & 99.3 &\textbf{100}&65.1&\textbf{72.2}&65.1&\textbf{72.2} \\
            \midrule
          \textbf{All}   &20192       & 99.2      & 99.3 &\textbf{99.9}&87.1&\textbf{88.8}&87.1&\textbf{88.8} \\  
    	\bottomrule
    	
        \end{tabular}
    }
    \vspace{-4pt}
    \caption{Fully supervised learning performance compared among RuleTakers (RT), \prover (PV) and \mymodel (PB) on test split of DU5 after training on training split of DU5, reported in varying depth.}
    \label{tab:full-supervised}
    \vspace{-10pt}
\end{table}

\subsection{Few-shot Learning}
\label{sec:few-shot}
We explore the few-shot learning ability of \mymodel against \prover by reducing training data size. For the sake of comparison, we follow the same setting in \cite{saha2020prover}, that is, randomly reserve 30k, 10k, 1k queries of overall 69762 training queries to train the model, denoted as ``RQ''.

It's worth noting that in the DU5 training dataset, several queries can be asked from a shared context. To better explore the ability when varying the amount of training data, we conduct another set of experiments, denoted as ``RC''. Specifically, we first randomly select context that appeared in the DU5 training dataset by a varying percentage, i.e., 10$\%$, 5$\%$, 1$\%$, and then reserve training samples where the query is asked from the selected context. 
Results of both ``RQ'' and ``RC'' are showed in Table \ref{tab:few-shot}.  

\begin{table}[t]
    \centering

    \scalebox{0.9}{
        \small
        \begin{tabular}{cccccccc}
            \toprule
            
            \multicolumn{2}{c}{\multirow{2}{*}{\tabincell{c}{Train Data}}}&
            \multicolumn{2}{c}{\textbf{QA}}& \multicolumn{2}{c}{\textbf{PA}}&\multicolumn{2}{c}{\textbf{FA}}\\
		\cmidrule{3-4}\cmidrule{5-6}\cmidrule{7-8}
            &&PV&PB& PV&PB&PV&PB \\
                
            \midrule
             &100\%         & 99.3   &\textbf{99.9} &87.1&\textbf{88.8}&87.1&\textbf{88.8}\\
            \midrule
            \multirow{3}{*}{\tabincell{c}{RC}}

            &10$\%$         &94.5      &\textbf{99.9}  &\textbf{63.6}&60.4&\textbf{63.3}&60.4 \\
            &5$\%$         &80.6       &\textbf{99.7}  &34.0&\textbf{44.2}&32.1&\textbf{44.2} \\
            &1$\%$         & 70.2     & \textbf{88.2} &20.0&\textbf{21.6}&15.1&\textbf{20.3} \\
            \midrule
            
          \multirow{3}{*}{\tabincell{c}{RQ}}
            
            &30k        & 97.8     & \textbf{99.9} &72.5&\textbf{86.8}&72.4&\textbf{86.8} \\
            &10k        &  87.1     &\textbf{99.9} &44.0&\textbf{72.4}&42.7&\textbf{72.3}  \\
            &1k         &  51.3     &\textbf{82.1}  &\textbf{28.0}&21.1&15.0&\textbf{18.4} \\

    	\bottomrule
    	
        \end{tabular}
    }
    \caption{Few-shot performance comparison among \prover and \mymodel on test split of DU5 after training on partial DU5 samples. (Two types of training samples, RC: queries from randomly reserved contexts; RQ: randomly reserved queries.)}
    
    \label{tab:few-shot}
\end{table}

Generally speaking, proof generation is harder to improve with increased training data, while QA performance improves rapidly by enlarging the training size. \mymodel widely defeats \prover on QA accuracy in each setting in Table \ref{tab:few-shot}. Surprisingly, \mymodel achieves 88.2$\%$ QA accuracy when training with only 700 samples (RC-1$\%$). Overall, \mymodel has a more stable ability for question answering when varying training data; however, \prover's QA accuracy drops sharply when lacking training data. This is because that \mymodel considers the joint distribution over all possible proofs and answers, and can better learn to reason over natural language statements. While as for proof accuracy, even if in some settings, \mymodel loses to \prover (RC-1$\%$), we will soon discover that \prover overfits to the small training data (Section \ref{sec:zero-shot} and \ref{sec:generalize}).

Another interesting observation is that the full accuracy is not always consistent with the proof accuracy in few-shot learning, which is different from the observation in Section \ref{sec:fully-sup}. Furthermore, we find that the gap between \textbf{PA} and \textbf{FA} when using \mymodel is much smaller than that of \prover. This is because \prover trains in a multi-task way, where the question answering module and proof generation module could make independent errors, especially when training data is not enough. But \mymodel can better utilize limited data to reason, which again verifies the effectiveness of \mymodel.

\subsection{Zero-shot Evaluation}
\label{sec:zero-shot}
Following previous work \cite{clark2020transformers,saha2020prover}, we evaluate the out-of-distribution (OOD) performance of \mymodel against baselines on six sub-datasets of Birds-Electricity. We conduct zero-shot experiments using DU5-trained models, which means that the model does not see any bird-domain or any electricity-domain samples during training. Results are showed in Table \ref{tab:zero-shot}. 

\begin{table}[t]
    \centering

    \small{
    \scalebox{0.88}{
        \begin{tabular}{p{0.3cm}p{0.3cm}ccccccc}
            \toprule
            \multirow{2}{*}{\tabincell{c}{Test}}& \multirow{2}{*}{\textbf{Cnt}}&\multicolumn{3}{c}{\textbf{QA}}& \multicolumn{2}{c}{\textbf{PA}}&\multicolumn{2}{c}{\textbf{FA}}\\
		\cmidrule{3-5}\cmidrule{6-7}\cmidrule{8-9}
            &&RT&PV&PB& PV&PB&PV&PB \\
                
            \midrule
            \textbf{B1}      &40      &97.5    &95.0 &\textbf{100.0}&92.5&\textbf{100.0}&92.5&\textbf{100.0} \\
            \textbf{B2}    &40     &  100    & 95.0 &\textbf{100.0}&95.0&\textbf{100.0}&95.0&\textbf{100.0}   \\
            \textbf{E1}   &162       &  96.9     &100 &\textbf{100.0}&95.1&\textbf{97.5}&95.1&\textbf{97.5}  \\
            \textbf{E2}   &180       & 98.3     & 100 &\textbf{100.0}&91.7&\textbf{93.3}&91.7&\textbf{93.3} \\
            \textbf{E3}   &624       &  91.8     & 89.7 &\textbf{98.2}&72.3&\textbf{79.3}&71.8&\textbf{79.3} \\
            \textbf{E4}   &4224       & 76.7      & 84.8 &\textbf{95.6}&\textbf{80.6}&77.7&\textbf{80.6}&77.7 \\
            \midrule
          \textbf{All}   &5270       & 80.1      & 86.5 &\textbf{96.3}&\textbf{80.7}&79.3&\textbf{80.5}&79.3 \\  
    	\bottomrule
    	
        \end{tabular}
    }
    }
      \vspace{-2pt}
    \caption{Zero-shot performance comparison among RuleTakers, \prover, and \mymodel on Birds-Electricity dataset after training on DU5.}
    
    \label{tab:zero-shot}
\end{table}

For QA accuracy, \mymodel outperforms \prover and RuleTakers obviously in all of sub-datasets. As for proof accuracy, \mymodel performs better when the depth of the out-of-domain sample $\leq3$, while there is a PA drop compared to \prover when testing on E4. This is a very interesting thing: superficially, proof accuracy drops for complicated unseen queries, but the QA accuracy for out-of-domain queries improves a lot (11 points on E4: $84.8\% \rightarrow 95.6\%$). We save it for future work to explore the portability of the proof and how an out-of-domain proof can help with question answering.

Moreover, we evaluate the zero-shot performance after few-shot learning. In Table \ref{tab:zero-after-few}, we report the results when testing on Birds-Electricity after training the model only on partial DU5 (RC-$k$ and RQ-$k$, described in \ref{sec:few-shot}) training partitions. As shown in Table \ref{tab:zero-after-few}, when testing zero-shot performance after few-shot learning, \mymodel is well ahead of \prover on QA accuracy. However, as for proof accuracy, \mymodel seems worth than \prover on the zero-shot test. Again we point out this amazing observation. This indicates that data from different domains might have different proof form. The well-learned proof from one domain might not be directly adopted to another, but, by training with \mymodel, the well-learned proof from one domain can help answer out-of-distribution queries. 

\begin{table}[t]
    \centering

    \scalebox{0.9}{
    \small
        \begin{tabular}{cccccccc}
            \toprule
            
            \multicolumn{2}{c}{\multirow{2}{*}{\tabincell{c}{Train Data}}}&
            \multicolumn{2}{c}{\textbf{QA}}& \multicolumn{2}{c}{\textbf{PA}}&\multicolumn{2}{c}{\textbf{FA}}\\
		\cmidrule{3-4}\cmidrule{5-6}\cmidrule{7-8}
            &&PV&PB& PV&PB&PV&PB \\
                
            \midrule
            \multicolumn{2}{c}{100$\%$}         & 86.5   &\textbf{96.3} &\textbf{80.7}&79.3&\textbf{80.5}&79.3\\
            \midrule
            \multirow{3}{*}{\tabincell{c}{RC}}

            &10$\%$         &71.2      &\textbf{99.9}  &\textbf{59.4}&55.4&\textbf{59.2}&55.4 \\
            &5$\%$         &59.4       &\textbf{99.5}  &55.0&\textbf{69.1}&46.6&\textbf{69.0} \\
            &1$\%$         & 47.1     & \textbf{60.6} &15.1&\textbf{34.6}&10.6&\textbf{24.4} \\
            \midrule
            
          \multirow{3}{*}{\tabincell{c}{RQ}}
            
            &30k        & 83.3     & \textbf{99.9} &76.79&\textbf{76.91}&76.72&\textbf{76.91}\\
            &10k        &  78.2     &\textbf{99.7} &54.3&\textbf{56.6}&54.3&\textbf{56.6}  \\
            &1k         &  50.4     &\textbf{51.3}  &\textbf{59.5} &34.6&\textbf{29.9}&17.3 \\

    	\bottomrule
    	
        \end{tabular}
    }
    \caption{Zero-shot performance comparison between \prover and \mymodel after few-shot learning. Test on Birds-Electricity after training on DU5 or partial DU5 (RC-$k$ and RQ-$k$) training partitions. }
    
    \label{tab:zero-after-few}
\end{table}

\subsection{Generalization Ability}
\paragraph{Generalize to Unseen Depth}
We conduct experiments to explore how well \mymodel can generate proofs and provide answers at depths unseen during training. Following \prover, we train the model on the training splits of DU0, DU1, DU2, and DU3, respectively, and test the QA performance and proof performance on the overall DU5 test set. As DU5 contains queries with higher depth than those seen during training, we can evaluate the model's ability when generalized to higher depth. 

\begin{table}[t]
    \centering
\small
\scalebox{0.9}{
        \begin{tabular}{cccccccc}
        
            \toprule
            \multirow{2}{*}{\tabincell{c}{Train\\ Data}}&\multicolumn{3}{c}{\textbf{QA}}& \multicolumn{2}{c}{\textbf{PA}}&\multicolumn{2}{c}{\textbf{FA}}\\
		\cmidrule{2-4}\cmidrule{5-6}\cmidrule{7-8}
            &RT&PV&PB& PV&PB&PV&PB \\
                
            \midrule
            \textbf{DU0}                &53.5 &\textbf{68.7}&56.9&44.4&\textbf{50.7}&\textbf{42.8}&41.3 \\
            \textbf{DU1}           &63.5 &73.7&\textbf{97.7}&63.8&\textbf{63.9}&61.9   &\textbf{63.9}\\
            \textbf{DU2}                &83.9 &89.6&\textbf{99.9}&72.6&\textbf{74.5}&72.3& \textbf{74.4} \\
            \textbf{DU3}               &98.9  &98.6&\textbf{99.9}&79.1&\textbf{83.2}&79.1 &\textbf{83.2}\\
            \midrule
          \textbf{DU5}               &99.2  &99.3&\textbf{99.9}&87.1&\textbf{88.8}&87.1&\textbf{88.8} \\  
    	\bottomrule
    	
        \end{tabular}
      }
    \caption{Performance comparison between RuleTakers, \prover, and \mymodel when testing on DU5 after training on DU0, DU1, DU2, DU3, respectively.}
    
    \label{tab:generalize-depth}
    
\end{table}

As shown in Table \ref{tab:generalize-depth}, \mymodel performs better than RuleTakers and \prover on all of QA/PA/FA performance when training on D1, D2 and D3, especially a significant improvement on QA performance. \mymodel shows a high and comparable QA performance when training only on depth=1 ($97.7\%$), which demonstrates \mymodel's superior generalization ability on depth. This means \mymodel can perfectly answer complicated queries using only simple training samples, which reduces the cost of constructing training data.

\paragraph{Generalize to Complex Language}
\label{sec:generalize}
We also evaluate the robustness of \mymodel when generalized to more diverse natural language. Following \cite{clark2020transformers, saha2020prover}, we train our model on the combined training partitions of DU3 and ParaRules, and then test on the ParaRules test partition. The results in Table \ref{tab:para-ori} show that \mymodel is more robust for human-like language.

\begin{table}[t]
    \centering
\scalebox{0.8}
{
        \begin{tabular}{p{0.2cm}p{0.6cm}ccccccc}
            \toprule
            \multirow{2}{*}{\tabincell{c}{D}}&\multirow{2}{*}{\tabincell{c}{Cnt}}&\multicolumn{3}{c}{\textbf{QA}}& \multicolumn{2}{c}{\textbf{PA}}&\multicolumn{2}{c}{\textbf{FA}}\\
		\cmidrule{3-5}\cmidrule{6-7}\cmidrule{8-9}
            &&RT&PV&PB& PV&PB&PV&PB \\
            \midrule
            \textbf{0}                &2968 &99.8&99.7&\textbf{99.8}&99.5&
            \textbf{99.5}&99.4&\textbf{99.4} \\
            \textbf{1}           &2406 &99.3&98.6&\textbf{99.7}&98.0&\textbf{98.0}&97.3&\textbf{98.0}   \\
            \textbf{2}                &1443 &98.2&98.2&\textbf{99.9}&88.9&\textbf{88.9}&88.7&\textbf{88.9}  \\
            \textbf{3}               &1036  &96.7&96.5&\textbf{99.8}&90.0&\textbf{90.1}&89.9&\textbf{90.1} \\
            \textbf{4}               &142  &90.1&88.0&\textbf{100}&76.1&\textbf{82.4}&76.1&\textbf{82.4} \\
            \midrule
          \textbf{All}               &8008  &98.8&98.4&\textbf{99.8}&95.4&\textbf{95.6}&95.1&\textbf{95.5} \\  
    	\bottomrule
        \end{tabular}
        }
    \vspace{-2pt}
    \caption{Performance comparison between \prover and \mymodel when testing on ParaRules test partitions after training on D3 + ParaRules training partitions.} 
    \label{tab:para-ori}
\end{table}

To better test the generalization ability for complex natural language, we train the model only on DU5 or partial DU5 (RC-$k$ and RQ-$k$, described in \ref{sec:few-shot}) training partitions and test on test split of ParaRules. This is a more convincing setup since the model will never see the human-like language but all templated language during training. Results are shown in Table \ref{tab:para-better}. When testing on ParaRules after only training on DU5, \mymodel outperforms \prover by nearly \textbf{30 points} on QA accuracy($53.6\% \rightarrow 82.8\%$). A similar trend is observed for training on RC-$k$ and RQ-$k$ datasets, where \mymodel improves the QA accuracy when generalized to human-like natural language. And the change for proof accuracy is not significant between \mymodel and \prover, which supports the observation in Section \ref{sec:few-shot} and \ref{sec:zero-shot}, that \mymodel improves QA performance by joint question-answering and proof-generation learning, but not necessarily improve the proof performance.

\begin{table}[t]
    \centering

    \small
        \begin{tabular}{cccccccc}
            \toprule
            
            \multicolumn{2}{c}{\multirow{2}{*}{\tabincell{c}{Train Data}}}&
            \multicolumn{2}{c}{\textbf{QA}}& \multicolumn{2}{c}{\textbf{PA}}&\multicolumn{2}{c}{\textbf{FA}}\\
		\cmidrule{3-4}\cmidrule{5-6}\cmidrule{7-8}
           &&PV&PB& PV&PB&PV&PB \\
                
            \midrule
            \multicolumn{2}{c}{100$\%$}         & 53.6   &\textbf{82.8} &40.0&\textbf{43.8}&38.4&\textbf{41.6}\\
            \midrule
            \multirow{3}{*}{\tabincell{c}{RC}}

            &10$\%$         &64.4      &\textbf{89.3}  &\textbf{42.0}&41.3&\textbf{41.0}&40.3 \\
            &5$\%$         &73.6       &\textbf{84.7}  &33.1&\textbf{36.5}&29.3&\textbf{35.1} \\
            &1$\%$         & \textbf{59.0}     & 56.3 &\textbf{30.4}&25.0&18.1&\textbf{18.3} \\
            \midrule
            
          \multirow{3}{*}{\tabincell{c}{RQ}}
            
            &30k        & 59.0     & \textbf{85.8} &38.6&\textbf{43.2}&37.5&\textbf{41.7} \\
            &10k        &  59.7     &\textbf{87.7} &41.7&\textbf{42.3}&40.3&\textbf{41.3}  \\
            &1k         &  51.4     &\textbf{56.3}  &\textbf{35.0} &25.0&\textbf{18.4}&16.5 \\
            
    	\bottomrule
    	
        \end{tabular}
      \vspace{-4pt}
    \caption{Performance comparison between \prover and \mymodel when testing on ParaRules test partitions after training on DU5 or partial DU5 (RC-$k$ and RQ-$k$) training partitions.} 
    
    \label{tab:para-better}
\end{table}

    

\begin{figure*}[t]
\centering
    \includegraphics[max size={0.9\textwidth}{0.9\textheight}]{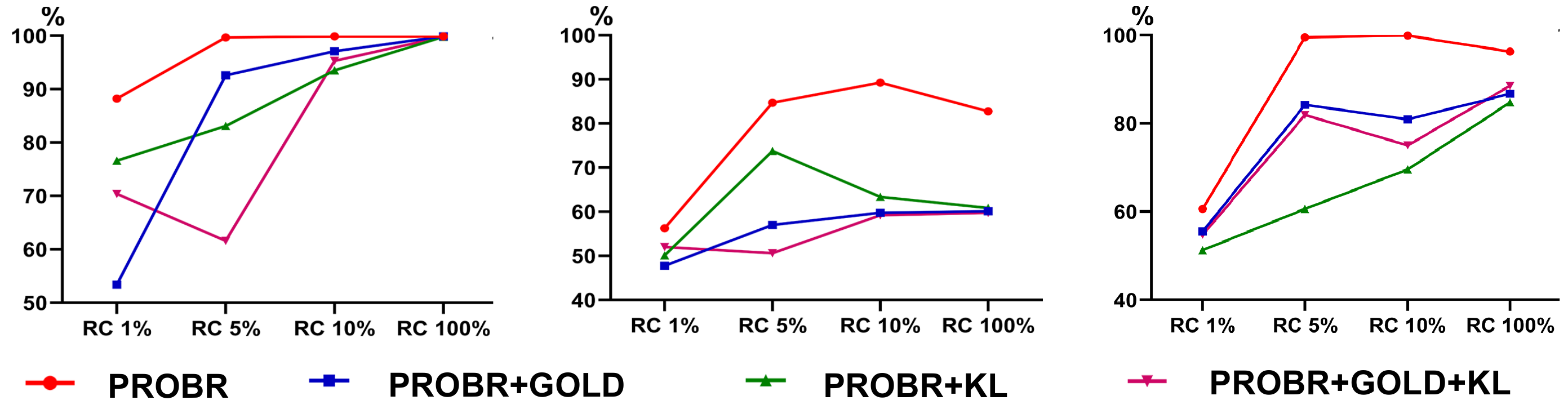}
	\caption{ 
	QA accuracy compared among \mymodel, \mymodel+ Gold, \mymodel+ KL, and \mymodel+ Gold + KL on DU5 test partition (left), on ParaRules test partitions (middle) and on Birds-Electricity dataset (right), after training on DU5 or partial DU5 (RC-$k$) training splits.}
    \label{fig:ablation}
\end{figure*}

\subsection{Ablation Studies}
\label{sec:abla}
We investigate the effect of training strategy and objective term $\mathcal{L}_\mathrm{kl}$ for our model. Specifically we compare \mymodel with the following three variants: 1) \textbf{\mymodel+ Gold}, that is, we replace predicted proofs with gold proofs when we optimize $\mathcal{L}_\mathrm{qa}$ during training. 2) \textbf{\mymodel+ KL}, that is, we add $\mathcal{L}_\mathrm{kl}$ between $q(Y)$ and $p_\mathrm{pseudo}(Y)$ during training. 3)\textbf{\mymodel+ Gold + KL} means both.
For \mymodel and above three variants, we first train on DU5 or partial DU5 (RC-$k$) training splits respectively, and Figure \ref{fig:ablation} reports the QA accuracy on test split of DU5 (left), ParaRules test partitions (middle) and Birds-Electricity (right). We observe that \mymodel always achieves the best QA accuracy on all of three test datasets (DU5, ParaRules, Birds-Electricity) after training on all of four datasets with varying size (RC-$1\%$, RC-$5\%$, RC-$10\%$, RC-$100\%$). And the other three model variants show inconsistent performance in different settings\footnote{For more details, please refer to the supplementary materials.}.

\section{Related Work}
\label{sec:related}


\paragraph{Text Reasoning over Formal Representation}
Early work employs a pipeline of methods that
converts free text into logic form first (semantic parsing),
and then uses formal logical reasoning \cite{musen1988brittleness}.
Due to the serious error propagation caused by semantic parsing \cite{zettlemoyer2012learning,berant2013semantic,berant2014semantic}, researchers focus on developing theorem provers 
by combining the symbolic techniques with the differentiable learning from neural networks \cite{reed2015neural,abdelaziz2020experimental,abboud2020learning},
such as NLProlog \cite{weber2019nlprolog},
SAT solving \cite{selsam2018learning} and
Neural programme \cite{neelakantan2015neural}.
To bypass this expensive and error-prone intermediate logical representation,
reasoning over natural language statements in an end-to-end manner is promising.

\paragraph{Text Reasoning over Natural Language}
Natural logic \cite{maccartney2009extended} focuses on semantic containment and monotonicity by incorporating semantic exclusion and implicativity.
Subsequently,
\citet{clark2020transformers} proposes to use a Transformer-based model to emulate deductive reasoning and achieves high accuracy on synthetically generated data.
\prover \cite{saha2020prover} points out that a reasoning system should not only answer queries but also generate a proof.
However, \prover adopts the multi-task learning framework in the training stage and cannot effectively capture the interactions
between question answering and proof generation.
Along this line,
we explore more powerful joint models to achieve deep reasoning.

\paragraph{QA and NLI}
There are bAbI \cite{weston2015towards}, QuaRTz \cite{tafjord2019quartz}, ROPES \cite{lin2019reasoning} and Hotpot QA \cite{yang2018hotpotqa} (QA datasets) involved in rule reasoning.
However, for those datasets,
implicit rules (i.e., which multi-hop chains are valid) need to be inferred from the training data.
In our task, the rules of reasoning are given in advance.
Compared with the Natural Language Inference \cite{maccartney2014natural}, 
our task can be regarded as its deductive subset.
In particular, NLI allows for unsupported inferences \cite{dagan2013recognizing}.

\section{Conclusion}
\label{sec:conclusion}
In this work, 
we propose \mymodel, a novel probabilistic graph reasoning framework for joint question answering and proof generation.
\mymodel defines a joint distribution over all possible answers and proofs,
which can directly characterize the interaction between answers and proofs.
Experiments prove the effectiveness of proposed \mymodel.

\section*{Acknowledgement}

The authors wish to thank the reviewers for their helpful
comments and suggestions,
the authors of RuleTakers for their datasets, 
and the authors of \prover for providing the source code.

\bibliography{main}
\bibliographystyle{acl_natbib}

\clearpage
\appendix

\twocolumn[
\begin{@twocolumnfalse}
    \section*{ \centering{ Supplementary Materials for \\ \emph{Probabilistic Graph Reasoning for Natural Proof Generation\\[30pt]}}}
\end{@twocolumnfalse}
]

\section{Experimental Environment}
\begin{table}[ht]
    \centering
    \begin{tabular}{lc}
        \toprule
        \textbf{Parameter} & \textbf{Value} \\
        \midrule
        Training Epochs & 30 \\
        Optimizer & AdamW \\
        Gradient Clipping & 1.0 \\
        Batch Size & 8 \\
        Dropout Rate & 0.1 \\
        Learning rate & 1e-5 \\
        Max Sequence Length & 300\\
        \bottomrule
    \end{tabular}
    \caption{Model configurations.}
    \label{tab:hyper-para}
\end{table}

Table \ref{tab:hyper-para} lists the default model configurations.
We produce \mymodel on 8 NVIDIA Tesla-V100 GPUs.
We implement \mymodel with PyTorch, using RoBERTa \cite{liu2019roberta} as pre-trained language model.

\section{Results on Development Set}
Table \ref{tab:full-supervised-dev} shows the results on development set under the fully supervised setting,
and the corresponding test set results are shown in Table \ref{tab:full-supervised}.
Overall, it is consistent with the findings of the test set results.
\mymodel achieves the best performance in three accuracy metrics (\textbf{QA}, \textbf{PA} and \textbf{FA}).
\begin{table}[ht]
    \small
    \centering
    \scalebox{0.9}{
        \begin{tabular}{p{0.2cm}p{0.6cm}p{0.4cm}cccccc}
            \toprule
            \multirow{2}{*}{\textbf{D}} & \multirow{2}{*}{\textbf{Cnt}}&\multicolumn{3}{c}{\textbf{QA}}& \multicolumn{2}{c}{\textbf{PA}}&\multicolumn{2}{c}{\textbf{FA}}\\
		\cmidrule{3-5}\cmidrule{6-7}\cmidrule{8-9}
            &&RT&PV&PB& PV&PB&PV&PB  \\
                
            \midrule
            \textbf{0}      &6299      &100    &100 &\textbf{100}&98.5&\textbf{98.7}&98.5&\textbf{98.7} \\
            \textbf{1}    &4434     &  98.4    & 98.8 &\textbf{100}&92.2&\textbf{93.6}&92.2&\textbf{93.6}   \\
            \textbf{2}   &2915       &  98.4     &99.2 &\textbf{99.0}&85.6&\textbf{87.3}&85.6&\textbf{87.3}  \\
            \textbf{3}   &2396       & 98.8      & 98.7 &\textbf{100}&82.8&\textbf{85.4}&82.8&\textbf{85.4} \\
            \textbf{4}   &2134       &  99.2     & 98.8 &\textbf{99.7}&76.9&\textbf{80.8}&76.9&\textbf{80.8} \\
            \textbf{5}   &2003       & 99.8      & 99.3 &\textbf{99.9}&67.4&\textbf{74.6}&67.4&\textbf{74.6} \\
            \midrule
          \textbf{All}   &20192       & 99.2      & 99.3 &\textbf{99.9}&88.0&\textbf{90.0}&88.0&\textbf{90.0} \\  
    	\bottomrule
    	
        \end{tabular}
    }
    \caption{Fully supervised learning performance compared among RuleTakers (RT), \prover (PV) and \mymodel (PB) on development split of DU5 after training on training split of DU5, reported in varying depth.}
    \label{tab:full-supervised-dev}
\end{table}

\begin{table}[t]
    \centering
\scalebox{0.9}{
        \begin{tabular}{ccccccc}
        
            \toprule
            \multirow{2}{*}{\tabincell{c}{Train\\ Data}}&\multicolumn{2}{c}{\textbf{QA}}& \multicolumn{2}{c}{\textbf{PA}}&\multicolumn{2}{c}{\textbf{FA}}\\
		\cmidrule{2-3}\cmidrule{4-5}\cmidrule{6-7}
            &PV&PB& PV&PB&PV&PB \\
                
            \midrule
            \textbf{DU0}                 &\textbf{68.3}&57.0&43.8&\textbf{50.7}&\textbf{42.3}&41.7 \\
            \textbf{DU1}            &73.2&\textbf{98.5}&63.9&\textbf{64.3}&61.8   &\textbf{64.3}\\
            \textbf{DU2}                 &89.3&\textbf{99.9}&72.6&\textbf{74.3}&72.3& \textbf{74.3} \\
            \textbf{DU3}                &98.3&\textbf{99.9}&79.4&\textbf{83.2}&79.4 &\textbf{83.2}\\
            \midrule
          \textbf{DU5}                &99.3&\textbf{99.9}&88.0&\textbf{90.0}&88.0&\textbf{90.0} \\  
    	\bottomrule
    	
        \end{tabular}
      }
    \caption{Performance comparison between \prover and \mymodel development split of DU5 after training on DU0, DU1, DU2, DU3, respectively.}
    
    \label{tab:generalize-depth-dev}
    
\end{table}

Table \ref{tab:generalize-depth-dev} lists the results on development set of DU5 after training on DU0, DU1, DU2, DU3, respective.
The corresponding test set resluts are shown in Table \ref{tab:generalize-depth}.
Comparing with Table \ref{tab:generalize-depth}, each number is very close (fluctuates within $1\%$ ) and similar conclusions can be drawn.


\section{Results of Ablation Studies}
Table \ref{tab:abla} lists the comparison results of \mymodel and the three variants of \mymodel (mentioned in Section \ref{sec:abla}).
Regarding the Table \ref{tab:abla},
we have three observations:
\begin{enumerate}[leftmargin=*, itemindent=1pc]
    \item In all ablation experiments, \mymodel achieved the best \textbf{QA} performance, demonstrating that \mymodel can capture critical information for question answering in a variety of settings.
    However, since some of the dataset are  artificially synthesized, it is difficult to guarantee that \mymodel will work in the real dataset as well.
    We leave it as future work.
    \item In some cases, variant d) (\textbf{\mymodel+ Gold + KL}) outperforms \mymodel in \textbf{PA} and \textbf{FA}.
    It shows the  potential advantages of the KL term.
    In the future, we will explore proof generation in a semi-supervised learning scenario through this KL term.
    \item When we compare the performance of the two models \mymodel and \textbf{\mymodel+ Gold}, 
    we can see that whether the predicted proof or the correct proof is used during training significantly affects the final performance. 
    Applying some heuristic strategies may give better results, such as scheduled sampling \cite{bengio2015scheduled}.
    We will try it in the future.
\end{enumerate}

\begin{table*}[]
\centering
\begin{tabular}{@{}l|lll|lll|lll@{}}
\toprule
\multirow{2}{*}{} & \multicolumn{3}{c|}{Test on DU5}                                                                                                                                                                                                   & \multicolumn{3}{c}{Test on ParaRules}                                                                                                                                                                                             & \multicolumn{3}{c}{Test on Birds-Electricity}                                                                                                                                                                                    \\ \cmidrule(l){2-10} 
                  & QA                                                                         & PA                                                                        & FA                                                                        & QA                                                                        & PA                                                                        & FA                                                                        & QA                                                                        & PA                                                                        & FA                                                                        \\ \midrule
RC-1\%            & \begin{tabular}[c]{@{}l@{}}a)\textbf{88.2}\\ b)53.4\\ c)76.6\\ d)70.4\end{tabular} & \begin{tabular}[c]{@{}l@{}}a)\textbf{21.6}\\ b)\textbf{21.6}\\ c)\textbf{21.6}\\ d)\textbf{21.6}\end{tabular} & \begin{tabular}[c]{@{}l@{}}a)20.3\\ b)\textbf{21.3}\\ c)14.8\\ d)20.0\end{tabular} & \begin{tabular}[c]{@{}l@{}}a)\textbf{56.3}\\ b)47.8\\ c)50.1\\ d)52.0\end{tabular} & \begin{tabular}[c]{@{}l@{}}a)\textbf{25.0}\\ b)\textbf{25.0}\\ c)\textbf{25.0}\\ d)\textbf{25.0}\end{tabular} & \begin{tabular}[c]{@{}l@{}}a)\textbf{18.3}\\ b)14.1\\ c)12.9\\ d)14.5\end{tabular} & \begin{tabular}[c]{@{}l@{}}a)\textbf{60.6}\\ b)55.5\\ c)51.3\\ d)54.8\end{tabular} & \begin{tabular}[c]{@{}l@{}}a)\textbf{34.6}\\ b)\textbf{34.6}\\ c)\textbf{34.6}\\ d)\textbf{34.6}\end{tabular} & \begin{tabular}[c]{@{}l@{}}a)\textbf{24.4}\\ b)17.4\\ c)17.6\\ d)19.9\end{tabular} \\ \midrule
RC-5\%            & \begin{tabular}[c]{@{}l@{}}a)\textbf{99.7}\\ b)92.6\\ c)83.1\\ d)61.6\end{tabular}  & \begin{tabular}[c]{@{}l@{}}a)44.2\\ b)43.7\\ c)31.4\\ d)\textbf{47.9}\end{tabular} & \begin{tabular}[c]{@{}l@{}}a)44.2\\ b)43.6\\ c)30.1\\ d)\textbf{45.9}\end{tabular} & \begin{tabular}[c]{@{}l@{}}a)\textbf{84.7}\\ b)57.0\\ c)73.8\\ d)50.6\end{tabular} & \begin{tabular}[c]{@{}l@{}}a)36.5\\ b)37.1\\ c)36.6\\ d)\textbf{38.9}\end{tabular} & \begin{tabular}[c]{@{}l@{}}a)35.1\\ b)35.4\\ c)27.6\\ d)\textbf{37.2}\end{tabular} & \begin{tabular}[c]{@{}l@{}}a)\textbf{99.5}\\ b)84.2\\ c)60.6\\ d)81.9\end{tabular} & \begin{tabular}[c]{@{}l@{}}a)69.1\\ b)62.0\\ c)45.3\\ d)\textbf{73.4}\end{tabular} & \begin{tabular}[c]{@{}l@{}}a)\textbf{69.0}\\ b)62.0\\ c)42.0\\ d)65.4\end{tabular} \\ \midrule
RC-10\%           & \begin{tabular}[c]{@{}l@{}}a)\textbf{99.9}\\ b)97.1\\ c)93.5\\ d)95.3\end{tabular}  & \begin{tabular}[c]{@{}l@{}}a)60.4\\ b)57.9\\ c)52.2\\ d)\textbf{69.3}\end{tabular} & \begin{tabular}[c]{@{}l@{}}a)60.4\\ b)57.9\\ c)51.8\\ d)\textbf{69.0}\end{tabular} & \begin{tabular}[c]{@{}l@{}}a)\textbf{89.3}\\ b)59.8\\ c)63.4\\ d)59.2\end{tabular} & \begin{tabular}[c]{@{}l@{}}a)\textbf{41.3}\\ b)40.0\\ c)40.7\\ d)37.5\end{tabular} & \begin{tabular}[c]{@{}l@{}}a)\textbf{40.3}\\ b)38.6\\ c)39.2\\ d)35.9\end{tabular} & \begin{tabular}[c]{@{}l@{}}a)\textbf{99.9}\\ b)80.9\\ c)69.6\\ d)75.0\end{tabular} & \begin{tabular}[c]{@{}l@{}}a)55.4\\ b)59.6\\ c)53.3\\ d)\textbf{66.2}\end{tabular} & \begin{tabular}[c]{@{}l@{}}a)55.4\\ b)59.6\\ c)53.2\\ d)\textbf{66.1}\end{tabular} \\ \midrule
RC-100\%          & \begin{tabular}[c]{@{}l@{}}a)\textbf{99.9}\\ b)\textbf{99.9}\\ c)99.8\\ d)99.8\end{tabular}  & \begin{tabular}[c]{@{}l@{}}a)88.8\\ b)88.7\\ c)89.1\\ d)\textbf{89.6}\end{tabular} & \begin{tabular}[c]{@{}l@{}}a)88.8\\ b)88.7\\ c)89.1\\ d)\textbf{89.6}\end{tabular} & \begin{tabular}[c]{@{}l@{}}a)\textbf{82.8}\\ b)60.1\\ c)60.9\\ d)59.8\end{tabular} & \begin{tabular}[c]{@{}l@{}}a)\textbf{43.8}\\ b)42.7\\ c)42.9\\ d)43.2\end{tabular} & \begin{tabular}[c]{@{}l@{}}a)\textbf{41.6}\\ b)39.8\\ c)41.2\\ d)41.5\end{tabular} & \begin{tabular}[c]{@{}l@{}}a)\textbf{96.3}\\ b)86.7\\ c)84.8\\ d)88.5\end{tabular} & \begin{tabular}[c]{@{}l@{}}a)79.3\\ b)\textbf{81.4}\\ c)78.6\\ d)80.9\end{tabular} & \begin{tabular}[c]{@{}l@{}}a)79.3\\ b)\textbf{81.3}\\ c)78.6\\ d)80.9\end{tabular} \\ \bottomrule
\end{tabular}
\caption{QA accuracy, proof accuracy and full accuracy compared among \mymodel, \mymodel+ Gold, \mymodel+ KL, and \mymodel+ Gold + KL on DU5 test partition (left), on ParaRules test partitions (middle) and on Birds-Electricity dataset (right), after training on DU5 or partial DU5 (RC-$k$) training splits, where a)--\textbf{\mymodel}, b)--\textbf{\mymodel+ Gold}, c)--\textbf{\mymodel+ KL}, d)--\textbf{\mymodel+ Gold + KL}.}
 \label{tab:abla}
\end{table*}

\end{document}